\documentclass{article}

 \usepackage[preprint,nonatbib]{nips_2018}

\usepackage[utf8]{inputenc} 
\usepackage[T1]{fontenc}    
\usepackage{hyperref}       
\usepackage{url}            
\usepackage{booktabs}       
\usepackage{amsfonts}       
\usepackage{nicefrac}       
\usepackage{microtype}      
\usepackage{graphicx}
\usepackage{amsmath,amssymb} 

\title{End-to-end driving simulation via angle branched network}

\author{
	Qing Wang\\
   School of Data and Computer Science\\ 
 Sun Yat-sen University \\
 Guangzhou, China\\
 \texttt{qingwang960710@outlook.com} \\
 \And
 Long Chen \\
 School of Data and Computer Science \\
 Sun Yat-sen University \\
 Guangzhou, China \\
 \texttt{chenl46@mail.sysu.edu.cn} \\
 \AND
 Wei Tian \\
 Institute of Measurement and Control Systems\\
 Karlsruhe Institute of Technology \\
 76131 Karlsruhe, Germany\\
 \texttt{wei.tian@kit.edu} \\
}

\begin{document}

\maketitle

\begin{abstract}
  Imitation learning for end-to-end autonomous driving has drawn attention from academic communities. Current methods either only use images as the input which is ambiguous when a car approaches an intersection, or use additional command information to navigate the vehicle but not automated enough. Focusing on making the vehicle drive along the given path, we propose a new navigation command that does not require human's participation and a novel model architecture called angle branched network. Both the new navigation command and the angle branched network are  easy to understand and effective. Besides, we find that not only segmentation information but also depth information can boost the performance of the driving model. We conduct experiments in a 3D urban simulator and both qualitative and quantitative evaluation results show the effectiveness of our model.
\end{abstract}

\section{Introduction}

Recently, using imitation learning to drive end-to-end caught researchers' attention again. Like human beings, imitation learning uses image as the input and the outputs are usually the steering wheel angle, acceleration and deceleration value. End-to-end driving means the model learns how to drive from expert's demonstration and it has been successfully applied to lane following \cite{Alvinn,Nvidia14} and off-road obstacle avoidance \cite{Lecun05}. The paper \cite{Cil} pointed out that using only image as the input is not enough to decide whether the vehicle should turn left or right, or go straight when reaching an intersection and proposed a method which is based on conditional imitation learning and use navigation command to handle this issue.

It is useful to think about what fators an end-to-end drving system should take into account. We analyse this problem from several points. First of all, navigation command is necessary for an end-to-end driving system to drive along the given path \cite{Cil}. Besides, from the perspective of human drivers, we know it is helpful to figure out what kinds of objects in front of the ego-vehicle. Xu et al. \cite{FcnDrive} proved that FCN (fully convolutional network) \cite{FCN} with privilege learning can boost the performance of the driving model. Furthermore, human drivers drive the car with attention. For example, when a human driver notices that the traffic light turns red, he will stop even though there are not vehicles or pedestrians ahead of the ego-vehicle. Xu et al. \cite{FcnDrive} also demonstrate that segmentation information can make the model pay attention to the right objects. However, when the car reaches an intersection, segmentation information may not be enough to let the car focus on right objects because the model can not decide where to see and where to go at this time without navigation command. Intuitively, branched architecture with navigation command, which means that different parts in the model process different navigation situations, may help. Moreover, human drivers can drive correctly because they can not only recognize what class the front object is but also infer the distance between the vehicle and the object. Inspired from the factors, we also use depth images and segmentation images as the input of the model and find that it can boost the driving performance. Moreover, autonoumous driving systems should be able to follow other vehicles. Intuitively, using sequential images instead of single image as the input may have such effect. Previous researches \cite{FcnDrive,NipsDrive,FastFcnDrive} demonstrate that LSTM (long short term memory) \cite{Lstm} can promote the driving performance. Finally, other information such as speed is also needed in an end-to-end driving system especially when the speed limit is considered.

Our task settings in this paper are a bit different comparing with task in \cite{Cil}. We consider the traffic light, other vehicles and the pedestrians but ignore the speed limit. Model's outputs include the steer angle and the throttle of the vehicle. We treat the throttle as a binary classification problem which means that the vehicle either stops or drives in a constant speed. Our contributions are as follows:

Firstly, we propose a new navigation command which is called as subgoal angle. To construct it, we divide a path into many uniform discrete points which are called as subgoal points. When driving along the path, what the vehicle's current subgoal is depends on where the vehicle travels along the path. Then we use car's current position and  current subgoal to calculate subgoal angle which is shown in Fig ~\ref{fig:subgoal_and_angle}. Section \ref{command} give a detailed description of the subgoal angle.

Secondly, in Section \ref{network}, we demonstrate a novel model  architecture using branched structure. In such architecture, neural network is used as the function approximator. The input of the model includes images from a single front-facing camera, car's current speed and our navigation command. The output of the model is the steering angle and the throttle. The navigation command plays two roles in our model. One is to act as detail information to better construct features for infering outputs, and the other is used as high level command to decide which branch of the output network to process features.

Thirdly, we find that depth information also has great influence on reducing the ego-vehicle's collision with other vehicles and pedestrians, which is really important in real world's application. This is illustrated in Section \ref{results}.

\section{Related work}
When constructing an end-to-end driving system, the first thing that we should consider is how we can test our model efficiently. Autonomous driving simulation environments have greatly boosted the study of end-to-end driving because they reduce research costs and time. Torcs \cite{Torcs} focuses on racing paths which are convenient to test alogrithms but not suitable for real autonomous driving applications. Euro Truck Simulator 2 and  Udacity's Self-Driving Car Simulator are good choices and adapt to test autonomous driving system in a highway simulation environment. Carla and Airsim \cite{Carla,Airsim} are ideal urban simulation environments and both have rich scenes as well as visual information. Besides, Carla's autopilot is very convenient for data collection and this is the main reason why we use Carla in our work.

In traditional methods for autonomous driving, the navigation is done by firstly using SLAM to build a 3D scene map and then planning a path in this map. Finally, a controlled approach is used to make the car drive along the planned path. Some of the recently methods including \cite{RuralDrive,DeepMindDrive} navigate the ego-vehicle without using fixed or detailed map. In \cite{RuralDrive}, the authors propose the local navigation goal, which is similar to our subgoal. The authors of \cite{DeepMindDrive} use some fixed landmarks to represent the goal and adopt reinforcemnt learning to navigate the agent. For the end-to-end driving, early works
\cite{Alvinn,Lecun05,Nvidia14,NipsDrive,FcnDrive,FastFcnDrive}  did not take the navigation into account which is not practical for real word applications. To our best knowledge, the paper reporting a conditional imitation learning method \cite{Cil} firstly considers this issue and integrates the navigation commmand into their driving model. Their navigation information includes three commands: turning left, going straight and turning right. In physical systems, they use a three-way switch on the remote control. This means that human should give the command information to the vehicle when driving, which is not automated enough for an autonomous driving system. Our method extracts the navigation command from the path and the vehicle's position, so it is more automated.

There are two trends in the study of training an end-to-end driving model. One is reinforcement learning. However, as far as we know, current driving approches based on reinforcement learning mostly belong to model-free reinforcement learning such as \cite{A3c,Ddpg} which learn how to drive by trial and error. These methods are difficult to apply to the real world because the training process is not safe. The other is imitation learning. Although imitation learning is easier to understand and implement, learning only from expert's demonstrations may lead to the problem that the policy is not able to recover from mistakes. Some promising methods raised in \cite{Dagger,Eril,Sbsp}
have mathematical guarantee. Other methods such as using left camera and right camera \cite{Nvidia14} and noise injection \cite{Cil} are empirically useful.

\section{Task definition}

We define the sequential observations $\{o_{t-k+1}, o_{t-k+2} \dots o_t\}$ as $o_{t-k+1:t}$ where $k$ is the number of the sequential obervations. Following the definition in \cite{Cil}, at the time step $t$, the controller receives sequential observations $o_{t-k+1:t}$, the measurment $m_t$ (e.g. speed) and the navigation command $c_t$, then takes an action $a_t$. To use the supervised learning method, we collect data $D =\bigl \{(o_{i-k+1:i},m_i,c_i,a_i)\bigr\}_{i=1}^n$ from the expert. In our task, we consider other vehicles and pedestrians. It is  necessary to set $k>1$ because the task of following other objects such as vehicles and pedestrians is  required. In our experiment, we set $k=4$. In other words, it can be seen as a partially observable markov decision process so we use sequential obervations to transform it into markov decision process. At time step $t$, the state information can be extracted from $o_{t-k+1:t}$, $m_t$ and $c_t$. In this way, there is an optimal policy $E$ that maps the observatons, the measurement, and the navigation command to the action, which can be described as $a_t = E(o_{t-k+1:t},m_t,c_t)$.

The problem lies in that we want to find a function $F$ parameterized by $\theta$ for $F\approx E$. The standard question can be formalized as

\begin{equation}
 \min\limits_{\theta}\sum\limits_{(o_{i-k+1:i},m_i,c_i,a_i)\in D}L\bigl(F(o_{i-k+1:i},m_i,c_i;\theta),a_i\bigr). 
\end{equation}

In our experiments, $o_{t-k+1:t}$, $m_t$ and $c_t$  represent sequential images from the front camera, the speed of the vehicle and the navigation command at time step $t$, respectively. Optimal actions $a_t$ are two dimensional vectors containing the steering angle and the throttle: $a_t = \langle
st_t,th_t\rangle$. We use $\hat{a_t}=\langle\hat{st}_t,\hat{th}_t\rangle$ as the result of  $F(o_{t-k+1:t},m_t,c_t)$. Mean square error is used for the steering angle and cross entropy error is used for the throttle. Therefore, the loss $L(\hat{a}_t,a_t)$ can be writen as 
\begin{equation}
  L\bigl( \hat{a}_t,a_t\bigr) = (1-\lambda)*MSE(\hat{st}_t,st_t) + \lambda*CrossEntropy({\hat{th}_t,th_t}).
\end{equation}
Here, $\lambda$ is the task weight between the two tasks and ranges from $0$ to $1$.

\section{Navigation command}
\label{command}

Our navigation command (i.e. subgoal angle) is calculated using the vehicle's current position and current subgoal. To get the current subgoal, we need to define the path properly.

\subsection{Path}
\label{path}

\begin{figure}
	\centering
	\includegraphics[width=0.3\linewidth]{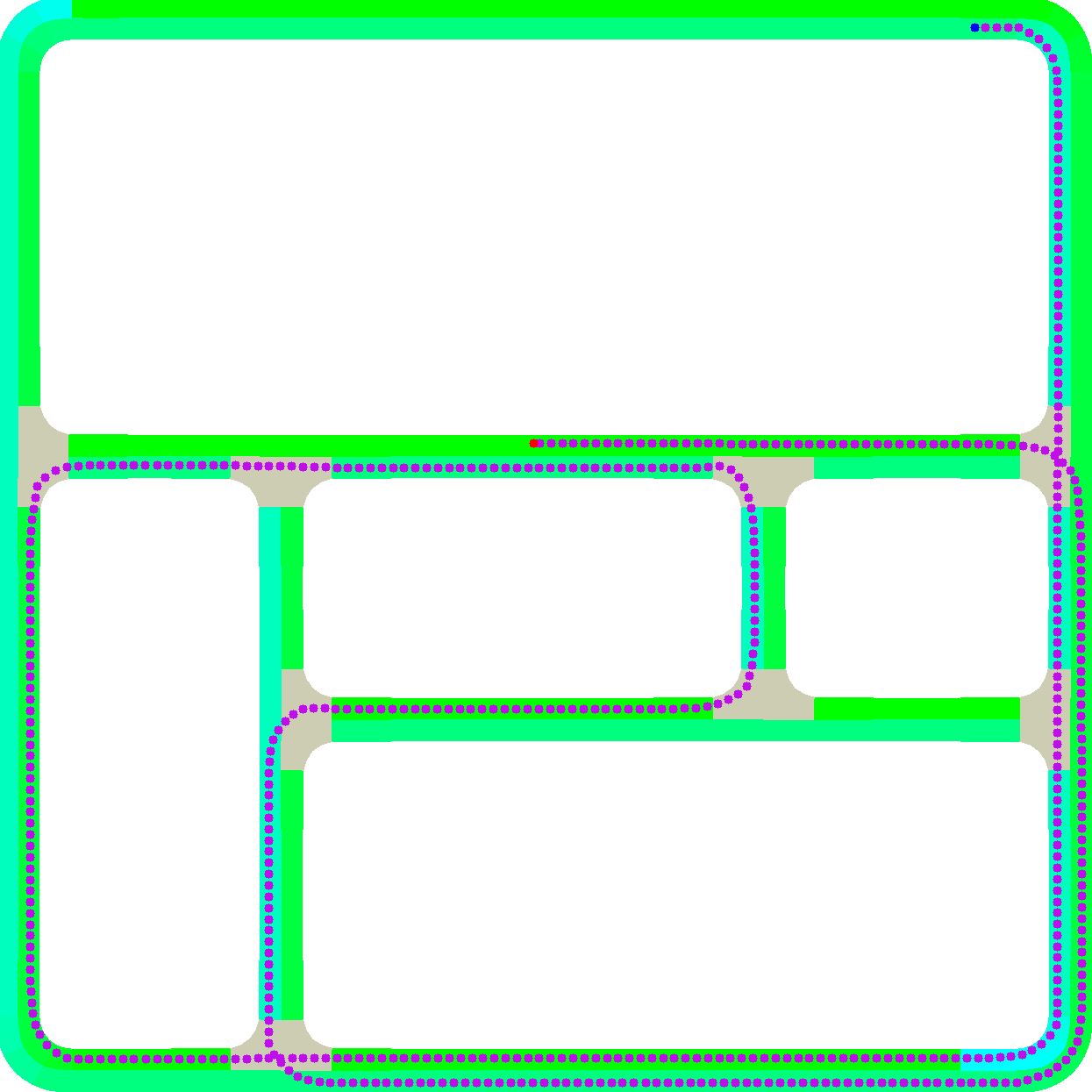}
	\caption{Visual representation of a path. The background is the map of town02 in Carla. The path is divided into many subgoal points. The distances between adjacent points are at least $x = 2$ meters in the simulator environment.}
	\label{fig:final_result}
\end{figure}

We divide a path into many uniform discrete subgoal points and guarantee that the distances between adjacent points are at least $x$ meters. In our experiments, we set $x=2$ by experience. An example can be seen in Fig ~\ref{fig:final_result}.

\subsection{ Subgoal direction}
\label{subgoal_direction}

It is of great importance to figure out how to define the subgoal. The first definition of a subgoal is often the shortest subgoal point away from the vehicle's current position. The problem lies in that the vehicle moves forward in a gradual process. Taking Fig ~\ref{fig:subgoal_and_angle} as an example, if the vehicle's current subgoal is the nearest subgoal point at time step $t+2$, it can't move along the path. Therefore, the subgoal of  the car is defined as the nearest subgoal point that is more than $d$ meters away from the car's current location. In our experiments, we find the performance is good enough when $d=3$. Let the vehicle's current position be $p_t$ and the current subgoal be $g_t$, the subgoal direction $sd_t$ is calculated by
\begin{equation}
sd_t = \frac{g_t-p_t}{{\lvert \lvert g_t-p_t \rvert \rvert_2}}.
\end{equation}

\subsection{ Subgoal angle}
\label{subgoal_angle}

\begin{figure}
	\centering
	\begin{minipage}{6cm}
		\centerline{\includegraphics[width=0.8\linewidth]{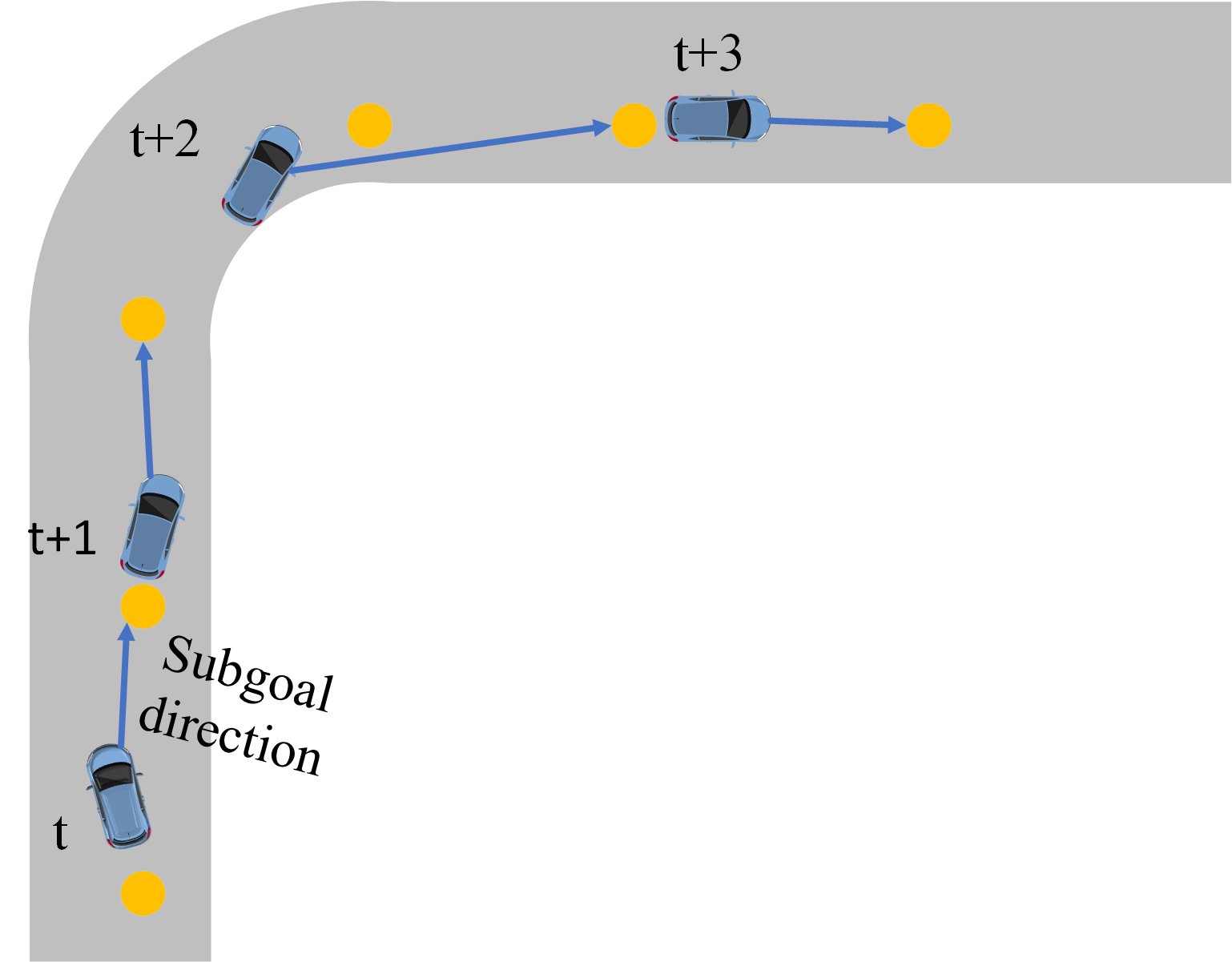}}
		\centerline{(a)}
		\label{fig:sg}
	\end{minipage}
	\begin{minipage}{6cm}
		\centerline{\includegraphics[width=1.2\linewidth]{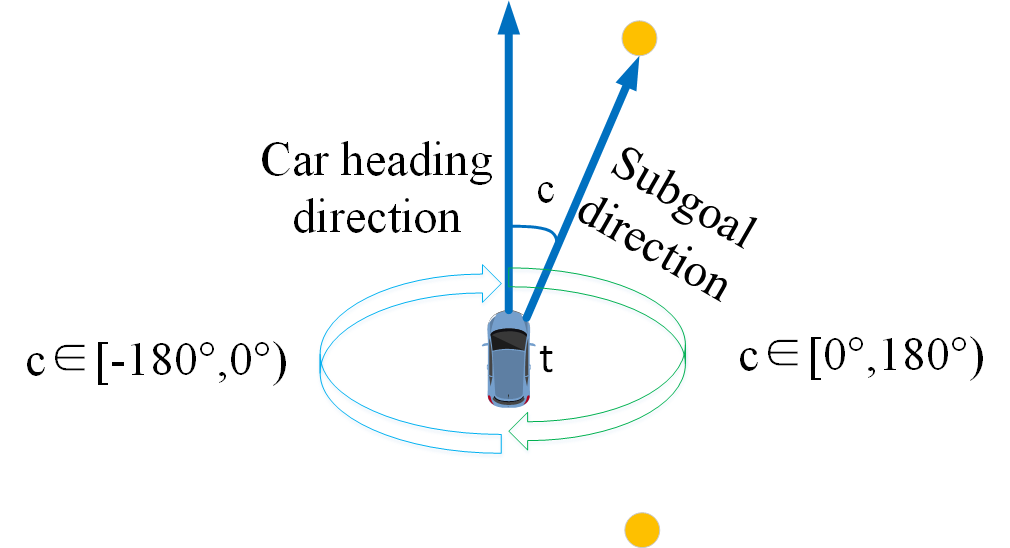}}
		\centerline{(b)}
		\label{fig:ang}
	\end{minipage}
	\caption{(a): An example of the subgoal direction. Yellow circles are subgoal points of a path. Letters around the cars indicate the time step and the subgoal direction points to the car's current subgoal; (b): An example of the subgoal angle. The subgoal angle is the angle between the car's heading direction and the subgoal direction.}
	\label{fig:subgoal_and_angle}
\end{figure}

If the car coordinate system is not used, subgoal direction is not enough to allow the vehicle to travel in the specified direction because the vehicle's heading direction is not taken into account. In this case, we define the subgoal angle (i.e. our navigation command) $c_t$ as the angle between the ego-vehicle's heading direction $h_t$ and the subgoal direction $sd_t$. In this way, the subgoal angle $c_t$ can be expressed by
\begin{align}
c_t = angle(sd_t,h_t)
\end{align}

and $angle(x,y)$ means calculating angle between the vector $x$ and the vector $y$. As shown in Fig ~\ref{fig:subgoal_and_angle}, we let the range of angles on the right side be between $0^o$ and $180^o$, and the range of angles on the left side be between $-180^o$ and $0^o$.

\section{Network architecture and details}
\label{network}

\begin{figure}
	\centering
	\includegraphics[height=0.38\linewidth]{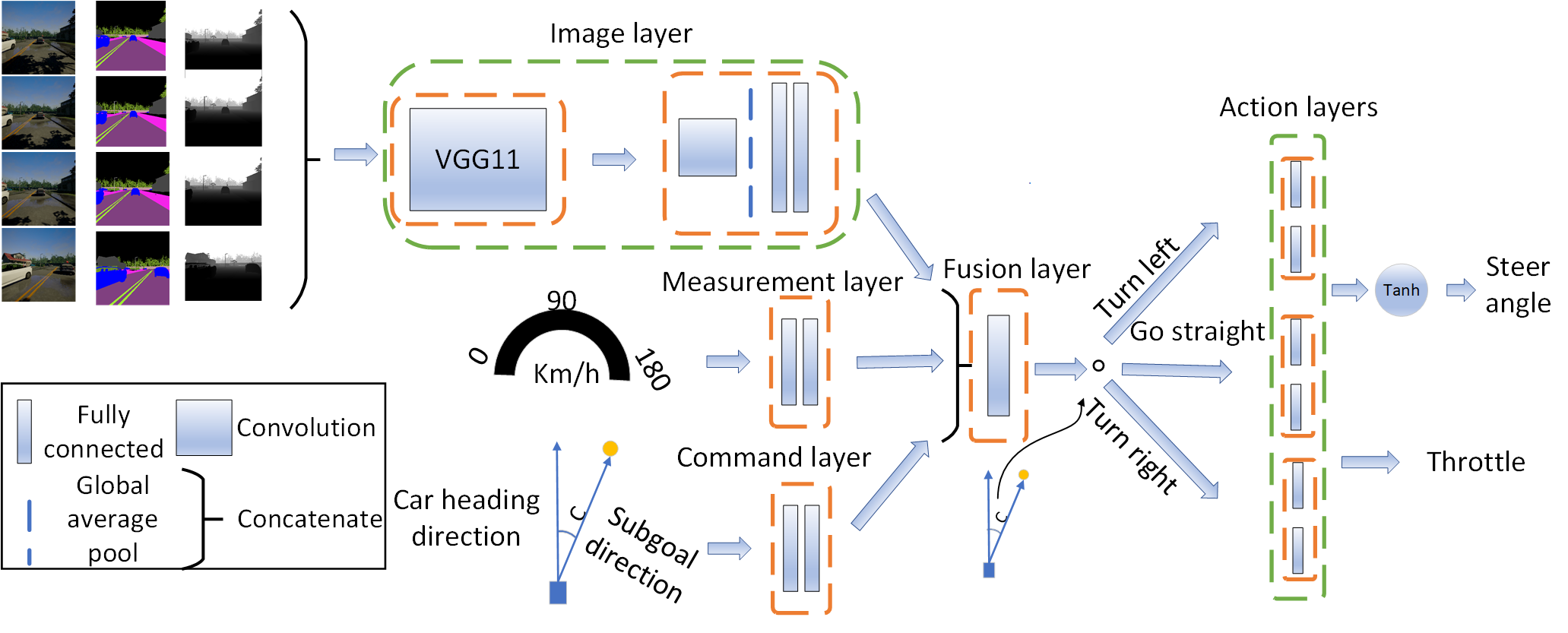}
	\caption{ Architecture of angle branched network.
	}
	\label{fig:angle_branched}
\end{figure}

We propose a new network architecture called angle branched network. As shown in Fig ~\ref{fig:angle_branched}, its inputs include the sequential images, the speed and the subgoal angle. We use first seven layers of VGG11 \cite{Vgg} pretrained on imagenet to process sequential images and change the channels of the first convolution layer of VGG11 to match the channels of the sequential images. Then we add one more convolution layer with 512 kernels to extract more useful features. A global average pooling layer and two fully connected layers with 512 hidden units are added behind the convolution layer transforming the image features into one 512-dimensional  vectors. The speed and the subgoal angle are processed with two fully connected layers (100 hidden units and 512 hidden units resepctively). Fusion layer is used to concatenate these three 512-dimensional  vectors into one 1536-dimensional  vectors and contains one fully connected layer with 100 hidden units. Finally, we discretize our subgoal angle into three parts, where $[-180^o,-10^o)$ means turning left, $[-10^o,10^o]$ means going straight, and $(10^o,180^o)$ means turning right. Depending on which part the subgoal angle is in, we use different action layers to process the features and make decisions.

Since driving scenes are complicated, it would be easier to learn robust features if we let different parts of network (i.e. the three action layers) to handle different situations (i.e. turn left, go straight, turn right). There are two key differences between our angle branched network and the branched network in \cite{Cil}. On the one hand, the subgoal angle is not only used as a switch to decide which action layer to process features but also served as the detailed navigation information to better fuse features. We can expect the network to work better because the value of the subgoal angle is usually positively related to the steer angle. On the other hand, the branched network in \cite{Cil} uses different fusion layers and action layers to fuse features from the image and the speed and then make decisions. However, we use only one fusion layer to fuse features from  sequential images, the speed and the subgoal angle but  different action layers to make decisions. In this way, we can reduce parameters and learn more robust features in the fusion layer.

\section{Experiments}

\subsection{Data collection}
Carla's autopilot is used to collect data including the observation, the measurement, and the action. From above information, we are able to construct the path and the navigation command. 

\subsubsection{Observation, measurement and action}
To enusre that the model can recover from mistakes, we also install the body of the left camera and the right camera on the ego-vehicle. Cameras are set on proper positions so that the body of the car is not shown in images. Depth and segmentation images are also captured in our experiment. It is enough to train a good policy by only using images from the three cameras, so we do not use other tricks like noise injection in \cite{Cil}.

In addition to images, we also collect the vehicle's own information including the position, the speed, the orientation (i.e. heading direction), the steering angle and the throttle etc. Notice that the steering angle is a floating-point type number between $-1$ and $1$, so the tanh layer is used after the action layer of our network architecture.

\subsubsection{Path and command}
To construct our paths, we arrange the vehicle's position in the chronological order. Then we pick up the location points chronologically and ensure that the distances between adjacent points are at least $x = 2$ meters. As we discussed in Section \ref{subgoal_direction}, we define the subgoal  of  the vehicle's each position as the nearest subgoal point that is more than $d = 3$ meters away. Given the current position and the subgoal, we can calculate the subgoal angle according to Fomula (3) and (4).

\subsubsection{Episode}
To get different paths, the data which we collect is in the form of episode. The frequency of collecting data ranges from 3 to 6 fps. Each episode contains 2000 frames and we get 350 episodes in town01 for training. For evaluation, we use the same way to get 50 paths in town02, but we just need the vehicle's position and ensure that each episode contains 4000 frames (i.e. each path is extracted from 4000 position points, and an example is shown in Fig ~\ref{fig:final_result}. There are 14 kinds of weather in Carla and each weather has an ID. To properly test the model's generalization, we use 7 kinds of weather (their weather IDs are 1, 3, 5, 7, 9, 12 and 14) in the training environment, and the rest kinds of weather (their weather IDs are 2, 4, 6, 8, 10, 11 and 13) are used for evaluation.


\subsection{Data preprocessing}
\begin{figure}
	\centering
	\begin{minipage}{6cm}
		\centerline{\includegraphics[width=6cm]{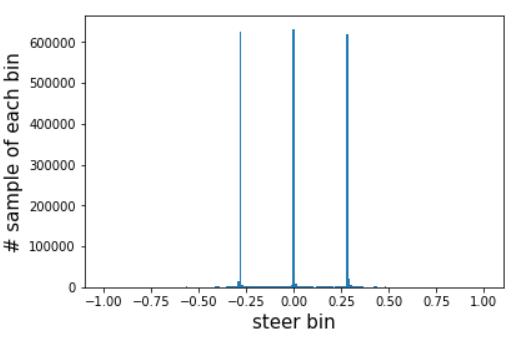}}
		\centerline{(a)}
	\end{minipage}
	\begin{minipage}{6cm}
		\centerline{\includegraphics[width=6cm]{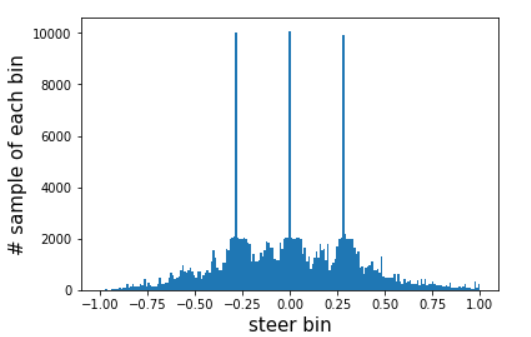}}
		\centerline{(b)}
	\end{minipage}
	\caption{Data distribution of the original data (a) and the preprocessing data (b). The steering angle ranging between -1 and 1 is divided into 199 bins. It is obvious that the data distribution before balancing is really unbalanced which is a big issue even though we use regression instead of classification when predicting steer angle.}
	\label{fig:data_balance}
\end{figure}

Data preprocessing is also an essential step. Most of the data we collect are straight driving data and we draw the distribution of the steer angle for visualization, which is shown in Fig ~\ref{fig:data_balance}. The policy trained from such unbalanced data will only focus on going straight. To solve this problem, we refer the idea of sampling. After dividing the steer angle into 199 bins, we randomly select some samples from each bin to ensure that the number of the samples from each bin will not surpass 2000. After such sampling process, there are few samples containing traffic light. Therefore, we select some samples containing traffic light and add them into the data set. Fig ~\ref{fig:data_balance} gives the comparison between the data distribution before and after preprocessing. The labels of the throttle are also unbalanced, so we set the class weight to inverse of the frequency of different classes. 

\subsection{Baseline}
Since we want to know the effect of our subgoal angle, we use the branched network in \cite{Cil} as our baseline. For fair comparison, we reimplement the branched network which can be done in our angle branched network by putting the switch before the fusion layer, using three fusion layers corresponding to three action layers and droping the command layer. Besides, to better test the effect of our angle branched network, we also use the command input network in \cite{Cil}. To distinguish our subgoal angle and their high level command, we call it angle input network. Angle input network can be implemented simply by droping the switch in the angle branched network and using just one action layer.

\subsection{Training}

Data augmentation, a very useful method, is widely used to increase the amount of data when training deep neural networks. In \cite{Cil}, data augmentation plays an important role in improving the network's performance. In our task, we find that data augmentation has little effect on our experiments, so we decide to ignore it. Xavier initialization \cite{Xavier} is adopted to initialize the weights of the convolution layer. We use the Adam \cite{Adam} optimization algorithm. The learning rate is set to 0.001 at the beginning and then it decreases as a factor 0.9. We use a batch size of 16. All models are implemented in Pytorch on a device with 32GB RAM and NVIDIA GeForce GTX Titan X GPU.

\subsection{Evaluation}
\subsubsection{Evaluation criteria}
We choose town02 in Carla as the evaluation environment and collect 50 paths. Each path is about 1.18 km. Two main evaluation criteria are defined in our experiments. One is the success rate which is the percentage of successful episodes, and the other is the normal-driving rate which is the ratio of the normal driving distance (collisions or veering outside the lane are non-normal driving) to the total driving distance of successful episodes. The success rate can measure whether the model is capable of fulfilling the task, and the normal-driving rate is used to measure how well the model fulfills the taks.

\subsubsection{Results}
\label{results}

\begin{figure}
	\centering
	\includegraphics[height=0.25\linewidth]{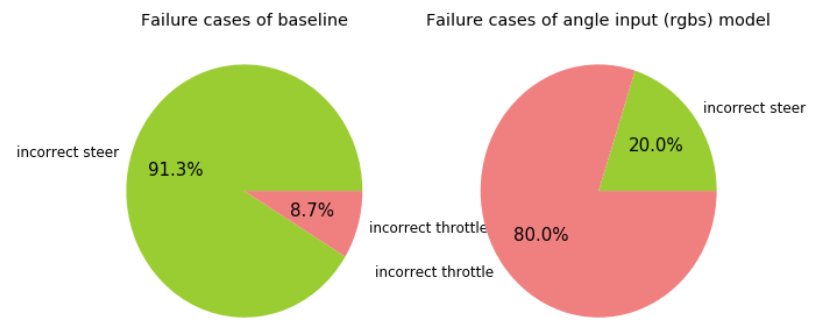}
	\caption{Comparision of failure cases of the baseline and the angle input (rgbs) model.
	}
	\label{fig:fail_case}
\end{figure}

\begin{figure}
	\centering
	\begin{minipage}{6cm}
		\centerline{\includegraphics[width=1.0\linewidth]{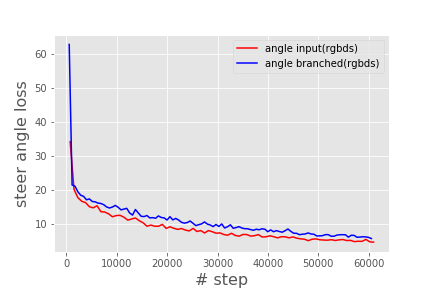}}
		\centerline{(a)}
		
	\end{minipage}
	\begin{minipage}{6cm}
		\centerline{\includegraphics[width=1.0\linewidth]{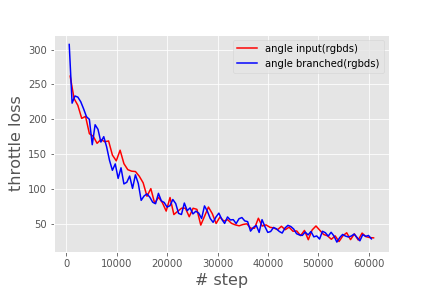}}
		\centerline{(b)}
		
	\end{minipage}
	\caption{Comparison of two losses between two models.}
	\label{fig:loss}
\end{figure}

We first demonstrate the effectiveness of the new navigation command (i.e. subgoal angle). When calculating the subgoal angle, we need to define the subgoal which is the nearest subgoal point of the path that is more than $d = 3$ meters. We use two combinations of different images as observation inputs including rgbs and rgbds (Here, rgb is the rgb image, d is the depth image and s represents the segmentation image). We also try using only rgb image as observation input but we find the performance is very bad so we ignore it here. For these two kinds of inputs, the pixel value of each channel of the rgb image and the depth image are normalized  to $[0,12]$ to make the numerical scale similar. From the result in Table~\ref{table:baseline_comparision}, three things are interesting. The first thing is the angle input (rgbs) model achieves the success rate of 92\% and the normal-driving rate of 99.5\%, which significantly boosts the performance of fulfilling the paths compared to baseline model. In experiment, we find most of the failure cases of the baseline are due to the incorrect prediction of steer angle while most of the failure cases of the angle input (rgbs) model are due to incorrect prediction of throttle. The comparison of situation of failure cases can be seen in Fig ~\ref{fig:fail_case}. It is not surprising for the reason that the value of the subgoal angle is always positively related to the steer angle thus easier to learn the mapping of the steer angle while the navigation commands in baseline model are just some high level commands and are not related to the steer angle. Besides, comparing the angle input (rgbds) model with the angle input (rgbs) model, it seems that depth information worsens the performance in terms of normal-driving rate. However, the angle input (rgbs) model always causes many collisions with other vehicle and pedestrian than the angle input (rgbds) model. We will show such comparision in section \ref{collision}. Finally, we find that the angle branched (rgbds) model performs best among all the models in terms of success rate and it also achieves high score in normal-driving rate. As Fig ~\ref{fig:loss} shows, the steer loss and throttle loss of the angle branched (rgbds) model decrease faster at the begining but converge to higher loss than the angle input (rgbds) model. But the angle branched (rgbds) model still gets higher score of success rate (94\% vs 92\%) and normal-driving rate (99.0\% vs 97.0\%) than the angle input (rgbds) model. We believe that it is because the angle branched architecture helps model learn more robust features and reduces the danger of overfitting.

\begin{table}
	\begin{center}
		\caption{Results of our models in Carla. We record the percentages of the successful episodes and the percentages of the normal driving distances. In both cases, higher value is better.}
		\label{table:baseline_comparision}
		\begin{tabular}{cccc}
			\toprule
			{\bf Model} & {\bf Size (MB)} & {\bf Success rate (\%)} & {\bf Normal-driving rate (\%)}\\
			\midrule
			Branched (rgbs, baseline) & 30.65 & 54 & 98.7\\
			Angle input (rgbs)& 30.27 & 92 & \bf{99.5}\\
			Angle input (rgbds) & 30.30 & 92 & 97.0\\
			Angle branched (rgbds) & 30.30 &\bf{94} & 99.0 \\
            \bottomrule
		\end{tabular}
	\end{center}
\end{table}

\subsubsection{Collision analysis}
\label{collision}

Since we consider the effect of other vehicles, pedestrians etc., the success rate and the normal-driving rate may be not enough to fully demonstrate the performance of driving model. Therefore, we also use some collision metrics including collision with pedestrians, vehicles and others (e.g. telegraph pole, sidewalk fence etc.). We define such collision metric as the average number of collision per kilometer of the successful episodes. Comparing the angle input (rgbs) model and the angle input (rgbds) model in Table~\ref{table:collision_comparision}, we know depth information can reduce the collision times in terms of collision-vehicle and collision-pedestrian which are directly related to human's lives in real word. Although the angle branched (rgbds) model performs worse than the angle input (rgbds) model, it is still better than the angle input (rgbs) model, which also shows the effectiveness of depth information. Therefore, depth information is useful in the end-to-end driving model.

\setlength{\tabcolsep}{3pt}
\begin{table}
	\begin{center}
		\caption{Collision analysis of all models. Collision-vehicle, collision-pedestrian, collision-others represent the collision with the vehicle, pedestrian and others. In all cases, lower value is better.}
		\label{table:collision_comparision}
		\begin{tabular}{cccc}
			\toprule
			{\bf Model} & {\bf Collision-vehicle} & {\bf Collision-pedestrian} & {\bf Collision-others}\\
			\midrule
			Branched (rgbs,baseline) & 0.63 & 0.53 & 1.19\\
			Angle input (rgbs)& 1.04 & 0.62 & \bf{0.72}\\
			Angle input (rgbds) &\bf{0.30} & \bf{0.41} & 1.15\\
			Angle branched (rgbds) & 0.42 & 0.52 & 1.05 \\
			\bottomrule
		\end{tabular}
	\end{center}
\end{table}

\section{Conclusion}
In this paper, we focus on the task of driving along the given path and present a new kind of navigation command which uses the vehicle's current position and current subgoal to calculate the subgoal angle for the end-to-end driving model. Based on this navigation command, we propose a novel model architecture using the branched structure. Besides, we also try different kinds of images as the observation input of the model including depth images as well as segmentation images and we find that depth information can reduce the collision with other vehicles and pedestrians, which is significant in real world. We demonstrate the effectiveness of our navigation command and the angle branched network by two main evaluation criteria and three collision metrics.

\bibliographystyle{abbrv}
\small
\bibliography{egbib}

\end{document}